\pdfoutput=1

\documentclass[11pt]{article}

\usepackage{acl}

\usepackage{times}
\usepackage{latexsym}

\usepackage[T1]{fontenc}

\usepackage[utf8]{inputenc}

\usepackage{microtype}

\usepackage{amsmath}

\usepackage{amsfonts}

\usepackage{soul}
\usepackage{tikz}
\usepackage{longtable}
\usepackage[export]{adjustbox}
\usepackage{subcaption}
\usepackage{makecell}

\usepackage{cleveref}
\usepackage{xcolor, color}
\usepackage{graphicx}
\usepackage{epstopdf}
\usepackage{multirow}
\usepackage{booktabs}
\usepackage{comment}
\usepackage{float}
\usepackage{stfloats} 
\usepackage{enumitem}
\usepackage{ctable} 
\usepackage{anyfontsize} 
\usepackage{pdfpages}
\usepackage{graphicx} 
\usepackage{epstopdf}
\usepackage{tabularx}

\usepackage{titlesec}
\usepackage{CJKutf8}

\title{Doctor Recommendation in Online Health Forums via Expertise Learning}

\author{Xiaoxin Lu$^{1}$\thanks{~~Equal contribution.~Yubo Zhang was supported by PolyU Undergraduate Research and Innovation Scheme (URIS).}~~~Yubo Zhang$^{1*}$~~~Jing Li$^{1}$\thanks{~~Jing Li is the corresponding author.}~~~Shi Zong$^2$ \\
$^1$Department of Computing, The Hong Kong Polytechnic University, HKSAR, China\\
$^2$Department of Computer Science and Technology, Nanjing University, Nanjing, China\\
$^{1}$\texttt{\{xiaoxin.lu, yubo.zhang\}@connect.polyu.hk} \\
$^{1}$\texttt{jing-amelia.li@polyu.edu.hk} \\
$^{2}$\texttt{szong@nju.edu.cn}
}

\begin{document}
\maketitle
\begin{CJK}{UTF8}{gbsn}
\begin{abstract}
Huge volumes of patient queries are daily generated on online health forums, rendering manual doctor allocation a labor-intensive task.
To better help patients, this paper studies a novel task of doctor recommendation to enable automatic pairing of a patient to a doctor with relevant expertise.
While most prior work in recommendation focuses on modeling target users from their past behavior, we can only rely on limited words in a query to infer a patient's needs for privacy reasons.
For doctor modeling, we study the joint effects of their profiles and previous dialogues with other patients and explore their interactions via self-learning.
The learned doctor embeddings are further employed to estimate their capabilities of handling a patient query with a multi-head attention mechanism.
For experiments, a large-scale dataset is collected from Chunyu Yisheng, a Chinese online health forum, where our model exhibits state-of-the-art results, outperforming baselines only considering profiles and past dialogues to characterize a doctor.\footnote{Our dataset and code are publicly available in:\quad \url{https://github.com/polyusmart/Doctor-Recommendation}}
\end{abstract}
\section{Introduction}

The growing popularity of health communities on social media has revolutionized the traditional doctor consultancy paradigm in a face-to-face manner. Massive amounts of patients are now turning to online health forums to seek professional help; meanwhile, popular healthcare platforms are able to recruit a large group of licensed doctors to provide online service \citep{liu2020meddg}.
In the COVID-19 crisis, the social distancing policies further flourish the use of these forums, where numerous patients would query diverse varieties of health problems every day \citep{gong2020internet}.

\begin{figure}[H]
\centering
\includegraphics[width=0.49\textwidth]{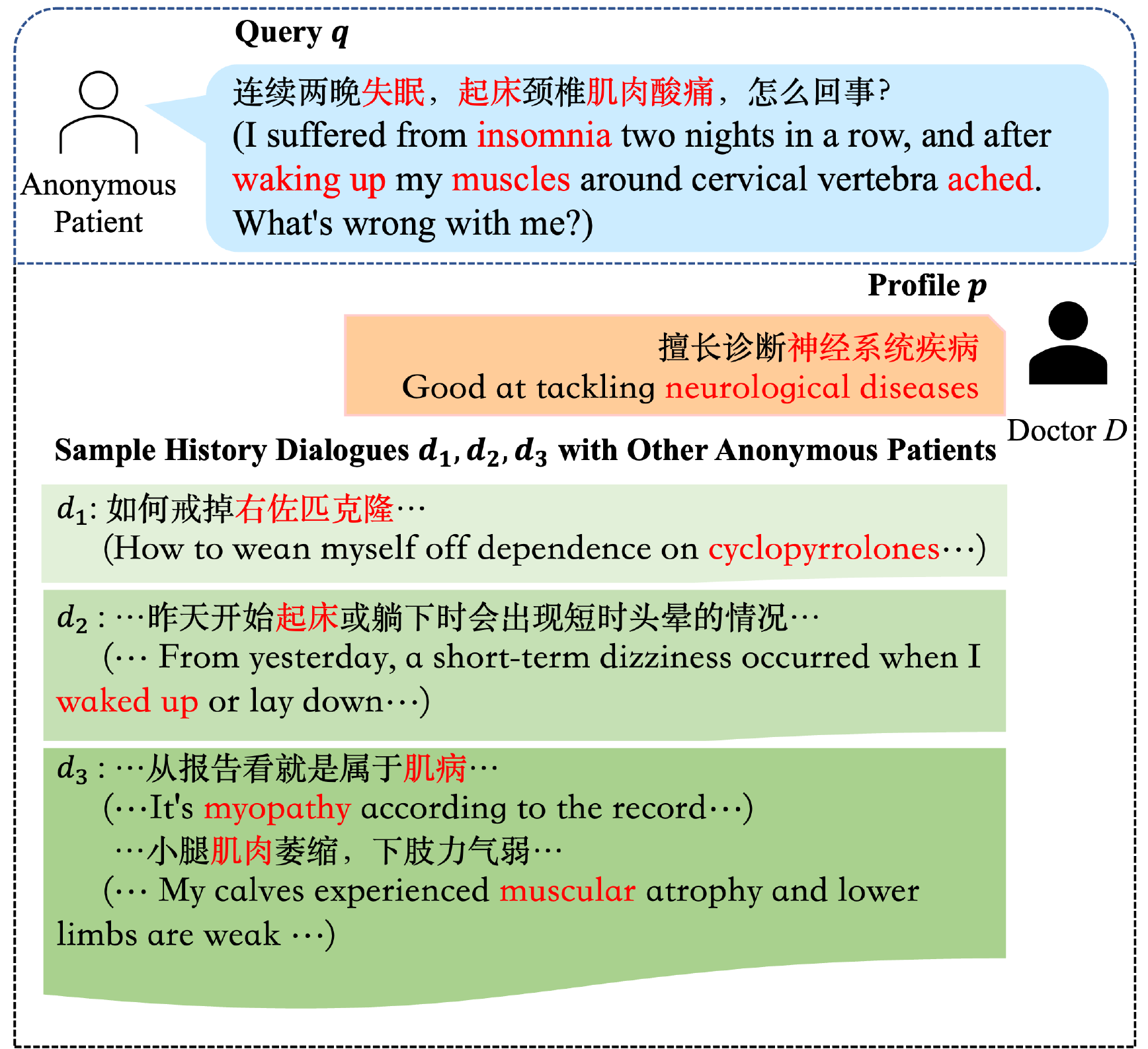}
\caption{The sample patient query $q$ on the top, followed by the profile of a sample doctor $D$ and three dialogues $D$ engaged before.
Salient words indicating patient needs and doctor expertise are 
in red.\protect\footnotemark
}
\label{fig:intro-case}
\end{figure}
\footnotetext{The original texts in our dataset are written in Chinese. We translated them into English in parentheses for reading.}

Nevertheless, in much practice \citep{cao2017online}, manual doctor allocations are adopted to handle each query, largely limiting the efficiency to help patients in sheer quantities and resulting in an extremely expensive process. Under this circumstance, how can we automate and speed up the pairing of patients to doctors who are able to offer the help?

In this paper, we present a novel task of \emph{doctor recommendation}, whose goal is to automatically figure out a patient's needs from their query on online health forums and recommend a doctor with relevant expertise to help.
The solution can not be trivially found from the mainstream recommendation approaches. 
It is because most recommender systems acquire the past behavior of target users (e.g., their purchase history) to capture their potential requirements \citep{wu-etal-2020-sentirec, Huang_Chen_Xia_Xu_Dai_Chen_Bo_Zhao_Huang_2021}; whereas our target users -- the patients -- should be anonymized to protect their privacy.
Language features consequently play a role in our task because only a few query words are accessible for models to make sense of how a patient feels and who can best help them.

To illustrate our task, Figure \ref{fig:intro-case} shows a patient's \textbf{query} $q$ concerning insomnia and muscle aches, where it is hard to infer the cause of such symptoms from the short text, not to mention to recommend a suitable doctor for problem-solving. It is hence crucial to explore the semantic relations between patient queries and doctor expertise for recommendation. To characterize a doctor's expertise, the modeling of their \textbf{profile} (describing what they are good at) provides a straightforward alternative. Nevertheless, the profiles are usually written in a professional language, while a patient tends to query with layman's terms. For instance, the doctor $D$ who later solved $q$'s problem is profiled with ``neurological diseases'', whose correlations with the symptom descriptions in $q$ are rather implicit. Therefore, we propose to adopt previous dialogues held by a doctor with other patients (henceforth \textbf{dialogues}) to narrow the gap of language styles between doctor profiles and patient queries. Take the history dialogues of $D$ in Figure \ref{fig:intro-case} as an example: the words therein like ``dizziness'', ``muscular atrophy'', and ``cyclopyrrolones'' (treatments for insomnia) are all helpful to bridge $D$'s expertise in neurological diseases with $q$'s symptoms.

To capture how a doctor's profile is related to their dialogue history, we first construct a self-learning task to predict whether a profile and a dialogue are from the same doctor.
It is designed to fine-tune a pre-trained BERT \citep{devlin2018bert} and align the profile writing and colloquial languages (used in patient queries and doctor responses) into the same semantic space to help model a doctor's expertise.
Profiles and dialogues are then coupled with the query embeddings to explore how likely a doctor is qualified to help the patient.
Here multi-head attention in aware of the doctor profile is put over the history dialogues to capture the essential content able to indicate a doctor's suitability from multiple aspects, e.g., the capabilities of $D$ in Figure \ref{fig:intro-case} to handle both ``insomnia'' and ``myopathy''. 
Such design reflects the intricate nature of health issues and would potentially allow the models to focus on the salient and relevant matters instead of being overwhelmed by the massive dialogues a doctor has engaged, which may concern diverse points.

In comparison to other NLP studies concerning health forum dialogues \citep{xu-etal-2019-doubletransfer,zeng-etal-2020-meddialog}, it is found that few of them attempt to spotlight doctors in these dialogues and examine how their expertise is reflected by what they say in these dialogues.
Different from them, we explore doctor expertise from their profiles and history dialogues in order to fit a doctor's qualification to a patient's requests, which would advance the so far limited progress of doctor expertise modeling with NLP.

To the best of our knowledge, \emph{we are the first to study doctor recommendation to automate the pairing of doctors and patients in online health forums, where the joint effects of doctor profiles and their previous interrogation dialogues are explored to learn what a doctor is good at and how they are able to help handle a patient's request. 
}

For experiments, we also gather a dataset with 119K patient-doctor dialogues involving 359 doctors from 14 departments from Chunyu Yisheng, a popular Chinese health forum.\footnote{\url{chunyuyisheng.com}}
The empirical results show that doctor profiles and dialogue history work together to well reflect a doctor's expertise and how they are able to help a patient.
In the main comparison, our model achieves state-of-the-art results (e.g., 0.616 by P@1), outperforming all baselines and ablations without employing self-supervised learning and multi-head attention.

Moreover, we quantify the effects of doctor profiles, history dialogues, and patient queries in recommendation and our model shows consistently superior performance in varying scenarios.
Furthermore, we probe into the model outputs to examine what our model learns with a discussion on multiple heads (in our attention map), a case study, and an error analysis, where the results reveal the potential of multi-head attention to capture various aspects of a doctor's expertise and point out the future direction to distinguish profile quality and leverage data augmentation and medical knowledge.
\section{Data Collection and Analysis}\label{sec:dataset}

Despite the previous contributions of large-scale data with doctor-patient dialogues \cite{zeng-etal-2020-meddialog}, 
we note some essential information for doctor modeling is missing, e.g., the profiles.
In this work, we present a new dataset to study the characterization of doctor expertise on health forums from both profiles and dialogue history.

\paragraph{Data Collection.} 

We developed an HTML crawler to obtain the data from Chunyu Yisheng, one of the biggest online health forums in China.
Then, seed dialogues involving 98 doctors were gathered from the ``Featured QA'' page.
To ensure doctor coverage in varying departments, we also collected doctors from the ``Find Doctors'' page for each department, which results in the 359 doctors in our dataset.
Finally, for each doctor, we crawled their ``Favorable Dialogues'' page and obtained the profile and history dialogues therein. 
All stop words were removed from each dialogue.

\paragraph{Data Analysis.}

The statistics of our dataset are reported in \Cref{tab:statistics}.
We observe that dialogues are in general much longer than profiles. We also observe that a doctor engages in over 300 dialogues on average. It indicates that rich information are contained in dialogues to learn doctor expertise, while presenting challenges to capture the essential content therein for effective doctor embedding.

\begin{table}[!h]
\small
\centering
\begin{tabular}{|lr|}
\hline
\# of dialogues & 119,128\\
\# of doctors  & 359 \\
\# of departments & 14\\
\hline
\hline
\# of tokens in vocabulary & 8,715\\
\hline
\hline
Avg. \# of dialogues per doctor & 331.83\\
Avg. \# of doctors per department & 25.64\\
\hline
\hline
Avg. \# of tokens in a query & 89.97 \\
Avg. \# of tokens in a dialogue & 534.28 \\
Avg. \# of tokens in a profile & 87.53\\
\hline
\end{tabular}
\caption{Data statistics. Each dialogue starts with a patient query and each doctor is associated with a profile.}
\label{tab:statistics}
\end{table}

\begin{figure}[htbp]
    \centering
    \includegraphics[width=0.49\textwidth]{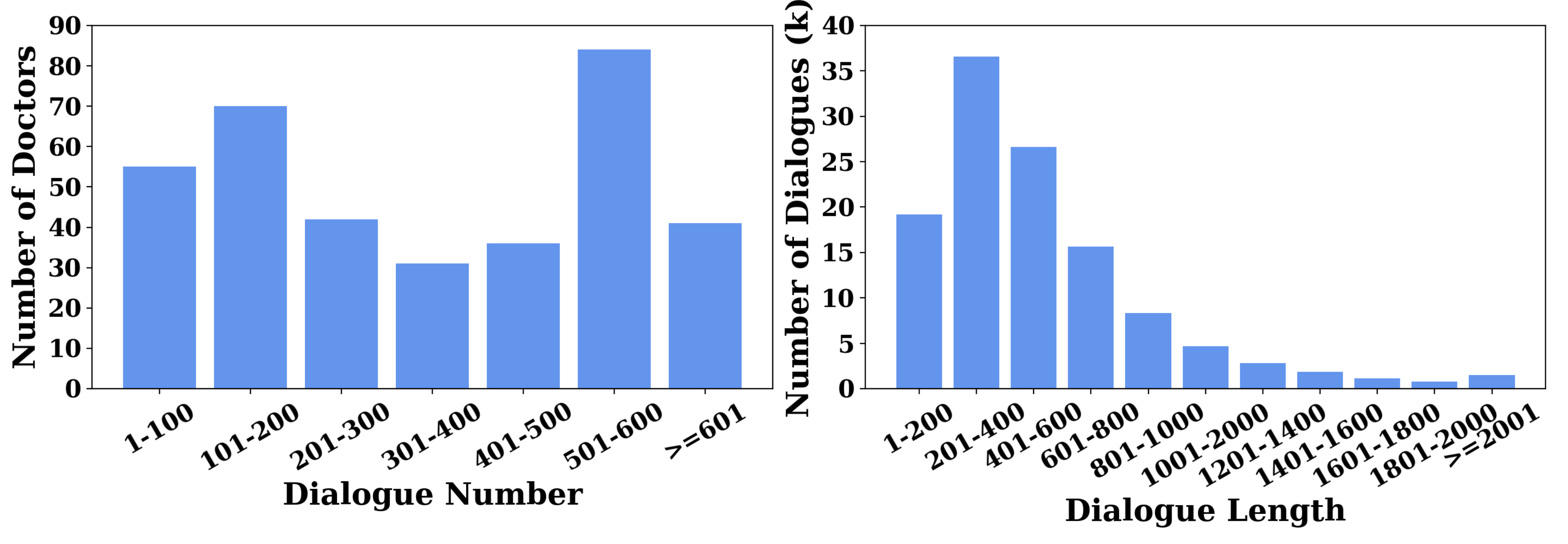}
    \caption{
    On the left subfigure, its y-axis shows the number of doctors and x-axis the dialogue number a doctor is involved in. For the right subfigure, the y-axis indicates the dialogue numbers in thousands (k) and x-axis the dialogue length in token number.
    }
    \label{fig:data-distribution}
\end{figure}

We further plot the distribution of dialogues a doctor engages and the dialogue length distribution in \Cref{fig:data-distribution}.
It is observed that doctors contribute diverse amounts of dialogues, which reflects the wide range of doctor expertise and qualifications in practice. 
Nonetheless, a large proportion of doctors are involved in over 100 dialogues while many dialogues are lengthy (with over 200 tokens).
We can hence envision a doctor's expertise may exhibit diverse aspects and dense information is available in history dialogues, whereas an effective mechanism should be adopted to capture salient content. 

We finally examine doctors' language styles by counting the number of medical terms based on THUOCL medical lexicon.\footnote{\url{github.com/thunlp/THUOCL/blob/master/data/THUOCL_medical.txt}}
Results show that medical terms take 30.13\% of tokens in doctor profiles, while the number is 7.83\% and 5.52\% for patient and doctor turns in dialogues, respectively.
It is probably because doctors tend to profile themselves with professional language while adopting layman's language to discuss with patients.
\section{Doctor Recommendation Framework}

\begin{figure}[H]
    \centering
    \includegraphics[width=0.45\textwidth]{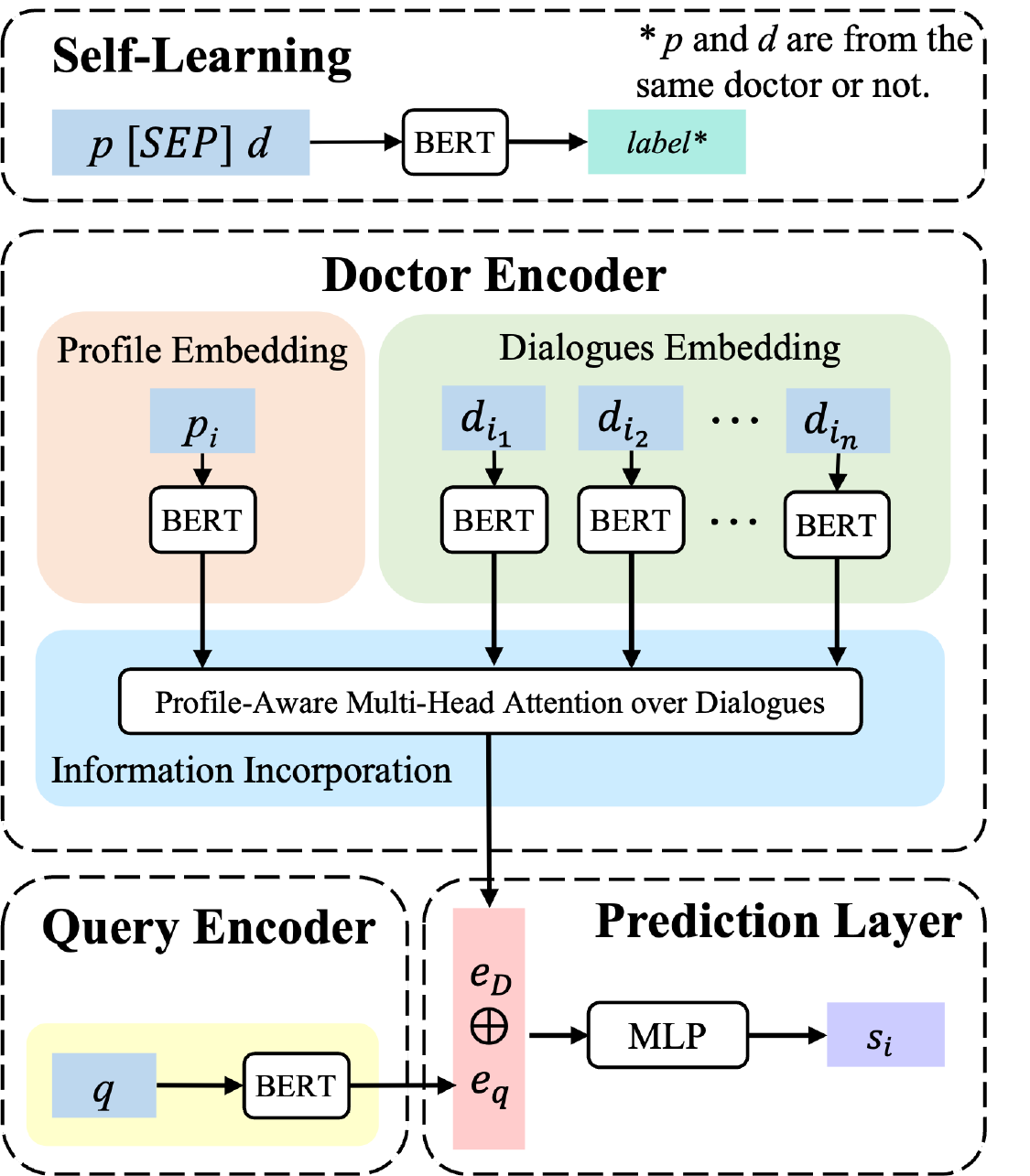}
    \caption{Overview of our framework.
    The doctor encoder first has its embedding layer (pre-trained BERT) fine-tuned via self-learning. 
    It then employs profile-aware multi-head attention over dialogues to explore doctor expertise and works with the query encoder (to capture patient needs) to pair doctors with queries. }
    \label{fig:framework}
\end{figure}

We now introduce the proposed framework for our doctor recommendation task (overviewed in \Cref{fig:framework}). 
It contains three modules: a query encoder that encodes patient needs from queries, 
a doctor encoder that encodes doctor expertise from profiles and dialogues,
and a prediction layer that couples above outputs for recommendation prediction.

\paragraph{Model's Input and Output.}

The input of our model is from three sources: a query $q$ from a patient, the profile $p_i$ of doctor $D_i$, and a collection of $D_i$'s history dialogues $\langle d_{i_1}, d_{i_2}, ..., d_{i_n}\rangle$ ($i_n$ denotes the number of dialogues $D_i$ previously engaged).
For each given query $q$, we first pair it with each doctor $D_i$ from a candidate pool of $m$ doctors
and output a matching score $s_i$ to reflect how likely $D_i$ owns the expertise to handle the request of $q$.
A recommendation is then made for $q$ by ranking all the doctor candidates based on these matching scores $s_i\,(i\in\{1, ..., m\})$.

\subsection{Doctor Encoder}
\label{ssec:doctor-encoder}

Here we introduce how we encode embeddings for a doctor $D$ to reflect their expertise, which starts with the embedding of their profile and dialogues.

\paragraph{Profile and Dialogue Embedding.} 

Built upon the success of pre-trained models for language representation learning, we employ a pre-trained BERT \cite{devlin2018bert} to encode the profile $p$ and obtain its rudimentary embedding $\mathbf{e}_p$. 
Likewise, for a dialogue $d$, we convert it into a token sequence via linking turns in  chronological order and encode its semantic features with BERT, which yields the dialogue embedding $\mathbf{e}_{d}$.

\paragraph{Self-Learning.} As analyzed in Section \ref{sec:dataset}, doctor profiles are usually written in a professional language while dialogue language tends to be in layman's styles.
To marry semantics of profiles and dialogues into the space, we design a self-learning task to predict whether a profile and a dialogue come from the same doctor, where random profile-doctor pairs are adopted as the negative samples. Then, the pre-trained BERT at doctor encoder's embedding layer is fine-tuned via tackling the self-learning task and shaping an initial understanding of how profiles are related to dialogues.

\paragraph{Multi-head Attention.} We have shown in Figure \ref{fig:data-distribution} that a doctor may engage in massive amounts of dialogues, where only part of them may be relevant with a query.
To allow models to attend to the salient information from the dense content provided by history dialogues, we put a profile-aware attention mechanism over dialogues. 
Here, multi-head attention is selected because of its capabilities in capturing multiple key points. 
It potentially reflects the complicated nature of doctor expertise, which in practice would exhibit multiple aspects.

Concretely, the profile embedding $\mathbf{e}_p$ is used to query and attend ${[\mathbf{e}_{d_1}, \mathbf{e}_{d_2}, …, \mathbf{e}_{d_n}]}^T$ (the dialogue embedding array) to both key and value argument:

\begin{align}
    \begin{cases}
    \text{Query}_{\text{att}} =\mathbf{e}_p,\\
    \text{Key}_{\text{att}}={[\mathbf{e}_{d_1},  \mathbf{e}_{d_2}, …, \mathbf{e}_{d_n}]}^T,\\ 
    \text{Value}_{\text{att}}={[\mathbf{e}_{d_1}, \mathbf{e}_{d_2}, …, \mathbf{e}_{d_n}]}^T.
    \end{cases}
\end{align}
For the $j$-th head, these three arguments are then respectively transformed through the neural perceptions with learnable weight matrices  $W^Q_j$, $W^K_j$, and $W^V_j$ ($Q$ for query, $K$ for key, and $V$ for value). 
Their outputs $\mathbf{Q}$, $\mathbf{K}$, and $\mathbf{V}$ jointly produce an intermediate doctor representation $\mathbf{h}_j$, which characterize a doctor's expertise from one perspective:
\begin{equation}
    \mathbf{h}_j = Att(QW^Q_j, KW^K_j, VW^V_j)
\end{equation}
where the $Att(\cdot)$ operation is defined as:
\begin{equation}
    Att(\mathbf{Q}, \mathbf{K}, \mathbf{V})=softmax(\frac{\mathbf{Q}\mathbf{K}^T}{\sqrt{dim}})\mathbf{V}
\end{equation}

\noindent Here $dim$ is the dimension of key and value. The scaling factor $\frac{1}{\sqrt{dim}}$ helps keep the softmax output away from regions with extremely small gradients.

Finally, to combine the learning results from multiple heads, outputs are concatenated altogether and transformed with a learnable matrix $W^O$ to obtain the final doctor embedding $\mathbf{e}_D$:
\begin{equation}
    \mathbf{e}_D = Concat(\mathbf{h}_1, \mathbf{h}_2, ..., \mathbf{h}_l)W^O
\end{equation}

\noindent Here $l$ denotes the number of heads. The doctor embedding $\mathbf{e}_D$, carrying features indicating the doctor expertise of $D$, will then be coupled with the query encoder results for recommendation, which will later be described in the coming section.

\subsection{Query Encoder and Prediction Layer}\label{ssec:query-encoder}

Then we describe how we measure the qualification of a doctor (embedded in $\mathbf{e}_D$) to handle a query $q$.

\paragraph{Query Embedding.}

For anonymous reasons, only the linguistic signals in a query are available to encode a patient's request.
Therefore, we adopt a similar strategy for the embedding of profiles and dialogues to customize the query encoder with a pre-trained BERT.
The learned feature is denoted as a query embedding $\mathbf{e}_q$ to represent patient needs.

\paragraph{Recommendation Prediction.} 

Given a pair of doctor $D$ and query $q$, the embedding results of doctor encoder $\mathbf{e}_D$ and query encoder $\mathbf{e}_q$ are coupled in the prediction layer for recommendation.
We adopt a MLP architecture to measure the matching score $s$ of the $D$-$q$ pair, which indicates the likelihood of doctor $D$ able to provide a suitable answer to query $q$ and is calculated as following:
\begin{equation}
    s = \sigma(W_{MLP}\cdot{Concat( \mathbf{e}_D,\mathbf{e}_q)}+b_{MLP})
\end{equation}
\noindent Here $\sigma$ denotes sigmoid activation function and $W_{MLP}$ (weights) and $b_{MLP}$ (bias) are trainable.

\subsection{Training Processes}

Our framework is based on the pre-trained BERT and then fine-tuned in the following two steps.
The first is to fine-tune the embedding layer of doctor encoder (as described in Section \ref{ssec:doctor-encoder}).
For the second, we fine-tune the entire framework by optimizing the weighted binary cross-entropy loss introduced in \citet{zeng-etal-2020-dynamic}:
\begin{equation}
\small
\label{eq:loss}
    L=-\sum_{(D,q)\in\tau}(\lambda\cdot{\hat{s}_{D,q}}\log({s_{D,q}})+(1-{\hat{s}_{D,q}})\log(1-{s_{D,q}}))
\end{equation}

\noindent
Here $\tau$ is the training set formed with doctor-query pairs and $\hat{s}_{D,q}$ denotes the binary ground-truth labels, with $1$ indicating $D$ later responded to $q$ while $0$ the opposite.
$\lambda>1$ balances the weights of positive and negative samples in model training, where the model would weigh more on positive $D$-$q$ pairs ($D$ indeed handled $q$) because negative samples may be less reliable and affected by many unpredictable factors, e.g., a doctor is too busy at some time.
Intuitively, this training objective encourages models to assign high matching scores $s_{D,q}$ to a doctor $D$ who actually helped $q$.

\section{Experimental Setup}
\label{sec:exp_setup}

We now describe the set up for our experiments.

\paragraph{Dataset Preprocessing and Split.}

To pre-process the data for non-neural models, we employed an open-source toolkit jieba for Chinese word segmentation.\footnote{\url{github.com/fxsjy/jieba}}
For neural models, texts were tokenized with the attached toolkit of MC-BERT, a pre-trained BERT for biomedical language understanding \cite{zhang2020conceptualized}, to be able to feed into BERT.\footnote{\url{github.com/alibaba-research/ChineseBLUE}} 
In the experiments, we maintained a vocabulary without stop words for dialogues' non-query turns while keeping them in queries and profiles, considering the high information density of the latter and colloquial styles of the former.

In terms of dataset split, 80\% dialogues were randomly selected from each doctor to form the training set. 
For the rest 20\% dialogues, we took their first turns (patient query) to measure recommendation and split the queries into two random halves, one for validation and the other for test.
In the training stage, we adopted negative sampling with a sampling ratio of 10 to speed up the process while for inference, the doctor ranking is conducted on the top 100 doctors handling the most queries.

\paragraph{Model Settings.}

As discussed above, the pre-trained MC-BERT was employed to encode the queries, profiles, and dialogues, whose parameters were first fine-tuned on the self-learning task, followed by a second fine-tuning step to tackle the doctor recommendation task with the other neural modules. 
The maximum input length of BERT is 512, and the dimension of all text embeddings from the output of MC-BERT is 768. The hyper-parameters are tuned on validation results and the following presents the settings.
The head number of multi-head attention is set to 6 and the tradeoff parameter $\lambda=5$ (Eq. \ref{eq:loss}) to weigh more on positive samples. The MLP at the output side contains one hidden layer in size 256. For training, we employ the Adam optimizer with an initial learning rate of 0.008 and batch size 256. The entire training procedure is 50 epochs, with early stop strategy adopted and the parameter sets result in the lowest validation loss used for test.

\paragraph{Baselines and Comparisons.}

We first consider weak baselines that rank doctors (1) randomly (henceforth \underline{\textsc{Random}}), (2) by the frequency of queries they handled measured on the training dialogues (henceforth \underline{\textsc{Frequency}}), (3) by referring to the doctors who responded to $K$ (in practice $K$ is set to 20) nearest patient queries in the semantic space (henceforth \underline{\textsc{KNN}}), (4) by the cosine similarity of profile and query embeddings yielded by the pre-trained MC-BERT (henceforth \underline{\textsc{Cos-Sim} (P+Q)}), and its counterpart matching dialogues and queries (henceforth \underline{\textsc{Cos-Sim} (D+Q)}). 
Then, a popular non-neural learning-to-rank baseline GBDT \cite{friedman2001greedy} with TF-IDF features is adopted (henceforth \underline{\textsc{GBDT}}).

For neural baselines, we compare with the MLP that simply matches query embeddings with profile embeddings (henceforth \underline{\textsc{MLP (P+Q)}}), with dialogue embeddings (henceforth \underline{\textsc{MLP (D+Q)}}), and with the average embeddings of profile and dialogue (henceforth \underline{MLP (P+D+Q)}).\footnote{We also test the alternative concatenates profile and dialogue embeddings, yet it results in very poor performance. 
A possible reason is the diverse styles of profile and dialogue languages and it is consistent with the observations from Table \ref{table:main-comp}, where concatenation operations tend to result in compromised performance. We will discuss more in Section \ref{ssec:comparison}.} 
We also consider Deep Structured Semantic Models (DSSM \cite{huang2013learning}), a popular latent semantic model for semantic matching. In this work, the original encoding bag-of-words module in DSSM is replaced with BERT. The query embeddings are matched with profile embeddings (henceforth \underline{\textsc{DSSM (BERT with P)}}) or the average embeddings of dialogues (henceforth \underline{\textsc{DSSM (BERT with D)}}).

To further examine the effects of our attention design for doctor modeling in recommendation, we attend a doctor's history dialogues in aware of their profile with two popular alternatives -- dot and concat attention \cite{luong-etal-2015-effective} (the former is henceforth referred to as \underline{\textsc{Dot-Att}} and the latter \underline{\textsc{Cat-Att}}).
They both went through a fine-tuning with the self-learning task before the training of recommendation to gain the initial view of how profiles and dialogues are related to each other. 
For comparison, we also experiment on our ablation based on multi-head attention without this self-learning step (henceforth \underline{\textsc{Mul-Att (w/o SL)}}).

At last, we examine the other two ablations that encode profiles only  with a multi-head self-attention (henceforth \underline{\textsc{Mul-Att (w/o D)}}) and its counterpart fed with dialogues only (henceforth \underline{\textsc{Mul-Att (w/o P)}}). 
The full model is henceforth named as \underline{\textsc{Mul-Att (full)}}.

For all models, we initialize them with three random seeds and average the results in three runs for the experimental report below.

\paragraph{Evaluation Metrics.} Following the common practice \cite{zeng-etal-2020-dynamic,zhang-etal-2021-howyoutagtweets}, the doctor recommendation results are evaluated with the popular information retrieval metrics: precision@$N$ (P@$N$), mean average precision (MAP), and ERR@$N$. In the experimental report, $N$ is set to 1 for P@$N$ and 5 for ERR@$N$, whereas similar trends hold for other possible numbers.

\section{Experimental Results}

In this section, we first present the main comparison results in Section \ref{ssec:comparison}.
Then, we quantify the model sensitivity to queries, profiles, and dialogues in varying lengths in Section \ref{ssec:quantitative}.
Finally, Section \ref{ssec:discussion} analyzes the effects of head number in validation performance, followed by a case study to interpret our superiority and error analysis to provide insights to future work.

\subsection{Main Comparison Results}
\label{ssec:comparison}

Table \ref{table:main-comp} reports the comparison results across different models. We draw the following observations.

First, it may require deep semantics to match doctor expertise with patient needs,  infeasible to rely on heuristic rules (e.g., frequency or similarity) or shallow features (e.g., TF-IDF) to well tackle the task.
Second, compared to profile, dialogues may better indicate how likely a doctor can help a patient, probably because of the richer content therein and the closer language style to a query (as analyzed in Section \ref{sec:dataset}).
Third, although the profiles and dialogues may potentially collaborate to better characterize a doctor (than the individual work), effective methods should be employed to couple their effects as their writings vary in the styles.

For models with multi-head attention, all of them yield better results than other attention counterparts.
This may imply the fact doctor expertise might be multi-faceted and multi-head attention works well to capture such feature.
We also notice a self multi-head attention over profile performs much worse than other ablations.
It is probably because profile content is very dense and may challenge multi-head attention in distinguishing various aspects therein. 

In comparison to \textsc{Mul-Att (w/o SL)}, \textsc{Mul-Att (w/o P)} (modeling doctors with dialogues only) and the results of our full model is almost twice better.
This again demonstrates the challenges present by the diverse wording patterns of profile and dialogues and the self-learning step to fine-tune pre-trained BERT would largely help in aligning them into the same semantic space.

\begin{table}[ht]
\small
\begin{center}
\begin{tabular}{|l|rrr|}
\hline
\textbf{Models} &P@1 &MAP &ERR@5\\

\hline
\hline

\underline{\textbf{Simple Baselines}}&&&\\

\textsc{Random}
&0.010 &0.052 &0.001\\

\textsc{Frequency}
&0.005 &0.032 &0.001\\

\textsc{KNN}
&0.082 &0.151 &0.008\\

\textsc{Cos-Sim (P+Q)}
&0.049 &0.122 &0.005\\

\textsc{Cos-Sim (D+Q)}
&0.056 &0.136 &0.006\\

\textsc{GBDT}
&0.018 &0.052 &0.002\\

\hline
\hline

\underline{\textbf{Neural Comparisons}}&&&\\

\textsc{MLP (P+Q)}
&0.164 &0.331 &0.018\\

\textsc{MLP (D+Q)}
&0.174 &0.341 &0.019\\

\textsc{MLP (P+D+Q)}
&0.153 &0.312 &0.017\\

\textsc{DSSM (BERT with D)}
&0.087 &0.182 &0.009\\

\textsc{DSSM (BERT with P)}
&0.151 &0.231 &0.012\\

\textsc{Dot-Att}
&0.219 &0.380 &0.021\\

\textsc{Cat-Att}
&0.167 &0.332 &0.018\\

\hline
\hline

\underline{\textbf{Our Ablations}}&&&\\


\textsc{Mul-Att (w/o SL)}
&0.309 &0.319 &0.019\\

\textsc{Mul-Att (w/o D)}
&0.198 &0.217 &0.013\\

\textsc{Mul-Att (w/o P)}
&0.521 &0.526 &0.033\\

\hline
\hline


\textsc{Mul-Att (full)}
&\textbf{0.616} &\textbf{0.620} &\textbf{0.039}\\
\hline
\end{tabular}
\end{center}
\caption{Results for doctor recommendation (averaged over queries). For all the metrics, the higher the better. 
Our model obtains the best results (in boldface) and significantly outperform others ($p$ < 0.02, paired $t$-test).}
\label{table:main-comp}
\end{table}

\subsection{Quantitative Analyses}
\label{ssec:quantitative}

In \Cref{ssec:comparison}, we have shown our model achieves a better performance compared to various baselines. In this section, we further quantify its performance in varying lengths of queries, dialogues, and profiles, and compare the full models' results with its two ablations \textsc{Mul-Att (w/o P)} and \textsc{(w/o SL)} -- the first and second runner-up in Table \ref{table:main-comp}. Afterwards, we provide the comparisons of model performance  across different medical departments to examine the scenarios where patients are able to know which department they should go to.

\paragraph{Sensitivity to Query Length.}

\begin{figure}[h!]
    \centering
    \includegraphics[width=0.48\textwidth]{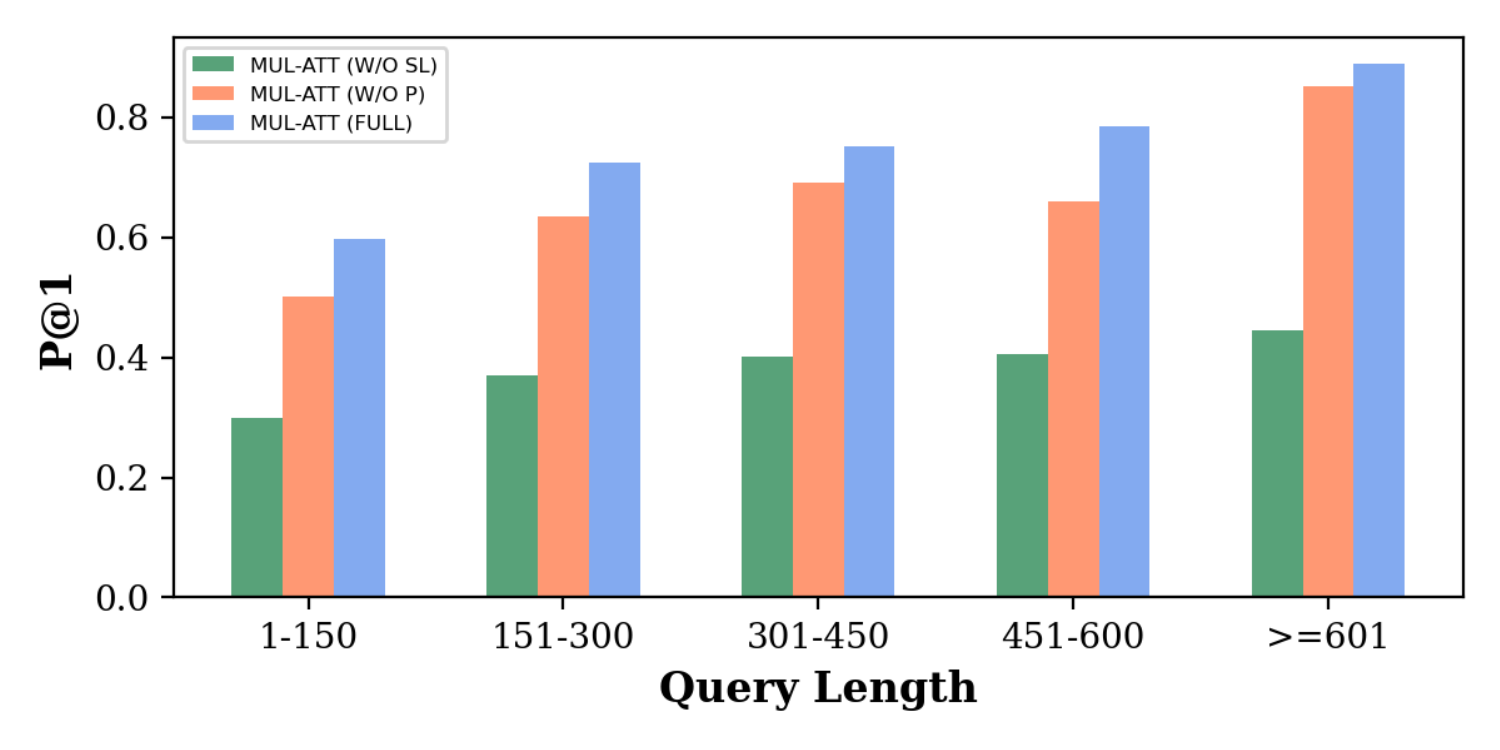}
    \caption{
    P@1 (y-axis) over varying query lengths (token number).
    For each group (x-axis), from left to right shows \textsc{Mul-Att (w/o SL)}, \textsc{(w/o P)}, and \textsc{(full).}
    }
    \label{fig:sens-query}
\end{figure}

Figure \ref{fig:sens-query} shows the P@1 over varying lengths of patient queries.
All models perform better for longer queries, owing to more content available to infer patient needs.
Besides, our full model consistently outperforms its two ablations while showing a relatively smaller performance gain for longer queries compared to \textsc{Mul-Att (w/o P)}.
A possible reason is: long queries may simplify the matching with doctors and dialogue content may be sufficient to handle recommendation, minoring the profile effects.

\begin{figure}[h!]
    \centering
    \includegraphics[width=0.48\textwidth]{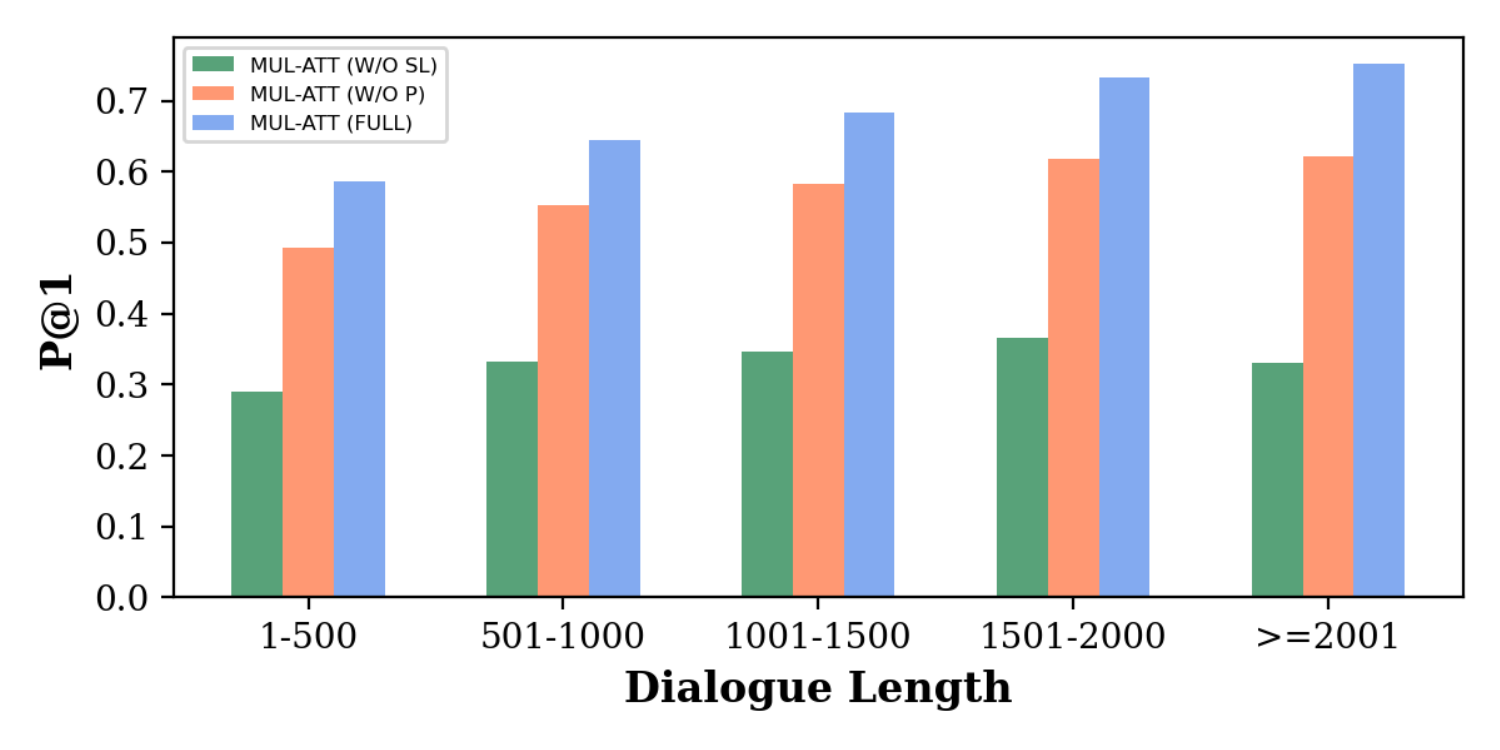}
    \caption{
    P@1 (y-axis) over varying dialogue lengths (token number).
    Bars in the x-axis are ordered the same as Figure \ref{fig:sens-query} and similar observations are drawn.
    }
    \label{fig:sens-dialog}
\end{figure}

\paragraph{Sensitivity to Dialogue Length.}
We then study the model sensitivity to the length of dialogues for doctor modeling and show the results in Figure \ref{fig:sens-dialog}.
Dialogue length exhibits similar effects to query length, possibly because they contribute homogeneous features to understand doctor-patient match. After all, other patients' queries are part of the dialogues and involved in learning doctor expertise. 

\paragraph{Sensitivity to Profile Length.}

\begin{figure}[h!]
    \centering
    \includegraphics[width=0.48\textwidth]{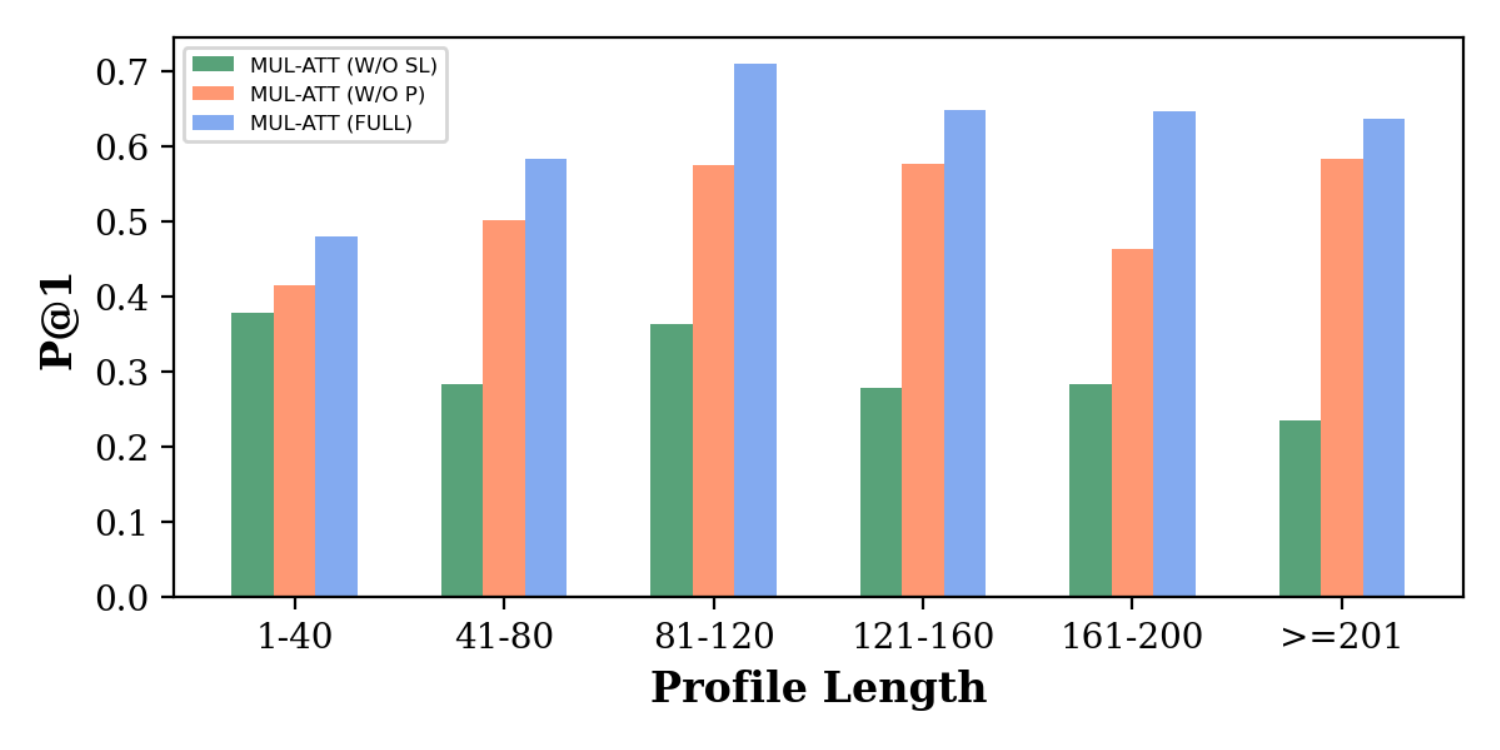}
    \caption{P@1 (y-axis) over varying profile lengths.
   We observe more complicated  effects compared to those from queries (Figure \ref{fig:sens-query}) or dialogues (Figure \ref{fig:sens-dialog}).
    }
    \label{fig:sens-profile}
\end{figure}

Furthermore, we quantify the profile length and display the models' P@1 in Figure \ref{fig:sens-profile}.
Here profile length exhibits different effects compared to query and dialogue length discussed above, where models suffer the performance drop for very long profiles, because of the potential noise therein hindering the collaboration with profiles and dialogues. 
Nevertheless, the self-learning step enables profiling language to blend in the colloquial embedding space of dialogues or queries, which hence presents more robust results.

\begin{figure*}[!h]
    \centering
    \includegraphics[width=0.9\linewidth]{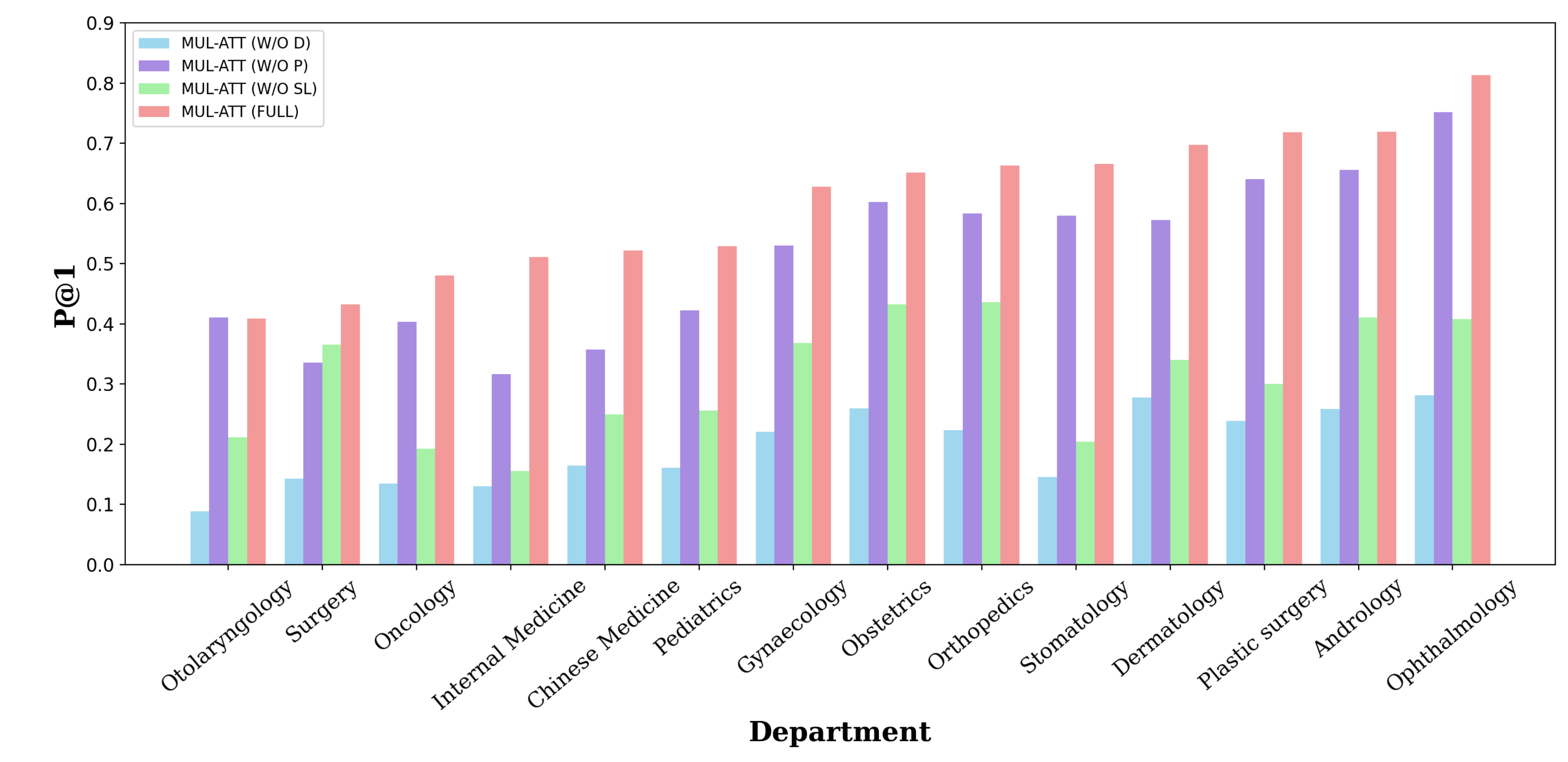}
    \caption{P@1 (y-axis) over all 14 departments (x-axis). 
    Our model achieves the best performance in 13 departments and obtains the comparable results to the best model for the left department of Otolaryngology.}
    \label{fig:p_score_per_dept}
    
\end{figure*}

\paragraph{Comparisons of Model Performance over Varying Departments.}

In the realistic practice, patients might have already known which department they should turn to before seeking help from doctors.
To better study doctor recommendation in this scenario, here we examine the model performance within different medical departments in our data. 
We select 4 models with highest P@1 scores in the main experiment (Table \ref{table:main-comp}) for comparison: \textsc{Mul-Att (w/o SL)}, \textsc{Mul-Att (w/o D)}, \textsc{Mul-Att (w/o P)}, and \textsc{Mul-Att (FULL)}. Their setups are described in Section \ref{sec:exp_setup}. 

Experimental results are shown in \Cref{fig:p_score_per_dept}. We observe for all 14 departments, our model has the best performance in 13 departments and achieves comparable results with the best model for the left department (otolaryngology). We also find all models exhibit varying performance when handling queries from different departments. It is related to departments' characteristics. For example, all models obtain low scores for Internal Medicine because of its significant overlap with others and the challenges to understand the needs from queries therein. 
Another factor is the imbalance of training data scale from each department. For instance, the training samples for Oncology, Surgery, Otolaryngology are much fewer than the average, resulting in the worse model performance on them.

\subsection{Further Discussions}
\label{ssec:discussion}

\paragraph{Analysis of Head Number.}

In Table \ref{table:main-comp}, multi-head attention shows the superiority to model doctors. We are hence interested in the effects of head numbers and vary them in validation set with the results shown in \Cref{tab:head_num}. It is seen that model performances first increase and then decrease, with 6 heads achieving the best performance. It indicates that head number reasonably affects model performance because it controls the granularity of aspects a model should capture to learn doctor expertise.

\begin{table}[h]\small

\begin{center}
\begin{tabular}{|l|rrr|}
\hline

\textbf{Head Number} &P@1 &MAP &ERR@5\\

\hline
\hline

\textsc{2}
&0.601 &0.605 &0.038\\

\textsc{4}
&0.609 &0.613 &0.038\\

\textsc{6 (OURS)}
&\textbf{0.616} &\textbf{0.620} &\textbf{0.039}\\

\textsc{8}
&0.564 &0.568 &0.035\\

\hline
\end{tabular}
\end{center}
\caption{The validation
results of our multi-head attention with different hyper-parameters in head number. 
}
\label{tab:head_num}
\end{table}

\paragraph{Case Study.}

To interpret what is learned by multi-head attention we take the example in Figure \ref{fig:intro-case} and analyze the attention map produced by 6 heads, where 4 of them attend to dialogue $d_3$ and the other 2 respectively highlights $d_1$ and $d_2$.
Recall that $d_1$, $d_2$, and $d_3$ each reflects a different aspects of doctor expertise.
To further probe into the attended content, we rank the words by the sum of attention weights assigned to a dialogue they occur in and show the top 5 medical terms in Table \ref{tab:case_att}. 
It is observed that the heads vary in their focusing point, while all related to the queried symptom of ``insomnia'' and  ``muscle ache'' and further contribute to a correct recommendation of a neurological expert.
This again demonstrates the intricacy of doctor expertise and the capabilities of multi-head attention to well reflect such essence.
More cases are shown in \Cref{appendix:case_study} to offer more insight of how our model recommends doctors.

\begin{table}[h!]\small
\newcommand{\tabincell}[2]{\begin{tabular}{@{}#1@{}}#2\end{tabular}}
\begin{center}
\begin{tabular}{|cl|}
\hline

\textbf{Head} $i$ &\textbf{Top 5 Keywords}\\

\hline
\hline

\textsc{1}
&\tabincell{l}{肌肉、神经、抽搐、无力、萎缩\\
(muscle, nerve, convulsion, weakness, atrophy)}\\
\hline

\textsc{2}
&\tabincell{l}{头晕，神经，头痛，内科，呕吐\\
(dizziness, nerve, headache, internal medicine,\\ sickness)}\\
\hline

\textsc{3}
&\tabincell{l}{神经，肌肉，酸痛，劳损，按摩\\
(nerve, muscle, ache, strain, massage)}\\
\hline

\textsc{4}
&\tabincell{l}{睡眠，焦虑，失眠，神经，右佐匹克隆\\
(sleep, anxiety, insomnia, nerve, Dexzopiclone)}\\
\hline

\textsc{5}
&\tabincell{l}{肌肉，颈部，头痛，恶心，颈椎\\
(muscle, neck, headache, sickness,\\
cervical vertebrae)}\\
\hline

\textsc{6}
&\tabincell{l}{神经，肌肉，颈部，酸痛，腰椎\\
(nerve, muscle, neck, ache, lumbar vertebrae)}\\

\hline
\end{tabular}
\end{center}
\caption{The top 5 medical terms attended by each head given the input sample in \Cref{fig:intro-case}. The medical terms are from the THUOCL lexicon used in Section \ref{sec:dataset}.}
\label{tab:case_att}
\end{table}

\paragraph{Error Analysis.}

We observe two major error types of our model, one resulting from doctor modeling and the other from the query.

For doctor modeling, we observe many errors come from the diverse quality of profiles.
As we have shown in Figure \ref{fig:sens-profile}, not all content from profiles is helpful.
For example, some doctors tend to profile themselves generally from experience (e.g., how many years they worked) instead of the specific expertise (what they are good at). 
Future work should concern how to further distinguish profile quality to learn doctor expertise.

In real world, some doctors are skilled comprehensively while others are more specialized. It causes the models tend to recommend the ``Jack of all trades'' rather than a more relevant doctor, as the former usually engaged in more dialogues and it is safer to choose them.
For example, in a query concerning ``continuous eye blinking'', the model recommends a doctor with 100 ``eyes''-related dialogues instead of the one specialized in ``Hordeolum'' and ``Conjunctivitis'' yet involved in only 30 dialogues.
To mitigate such bias, it would be interesting to employ data augmentation \cite{zhang-etal-2020-parallel} to ``enrich'' the history for doctors handling relatively fewer queries.

In terms of queries, many patients are observed to describe their symptoms with minutiae rather than focusing on the key points.
So the model, lacking professional knowledge, may consequently be trapped with these unimportant details. 
For instance, a patient queried a ``pimple'' on the ``eyelid''; the model wrongly attends to ``eyelid'' thus recommends an ophthalmologist but not a dermatologist to solve the ``pimple'' problem.
A future direction to tackle this issue is to exploit knowledge from medical domains \cite{liu2020k} to allow a better understanding of patient needs.

\section{Related Work}

Our work is in the research line of recommender systems widely studied because of their practical value in industry \citep{Huang_Chen_Xia_Xu_Dai_Chen_Bo_Zhao_Huang_2021}. 
For example, previous work explores users' chatting history to recommend conversations \cite{zeng-etal-2018-microblog, zeng-etal-2020-dynamic} and hashtags \cite{li-etal-2016-hashtag,zhang-etal-2021-howyoutagtweets}, browsing history to recommend news \cite{wu-etal-2019-neural-news, qi-etal-2021-pp}, and purchase history to recommend products \cite{guo-etal-2020-deep}. 
In contrast to most recommendation studies focusing on exploiting target users' personal interest modeling from their history behavior, our work largely relies on wordings of a short query to figure out what is needed by a target user (patient) because they are anonymous for privacy concern.

Within several branches of recommendation research, our task is by concept similar to expert recommendation for question answering \cite{wang2018survey,NIKZADKHASMAKHI2019126}.
In this field, many previous studies encode expertise knowledge in diverse streams, such as software engineering \cite{bhat2018expert}, social activities \cite{app11167681}, etc.
Nevertheless, few of them attempt to model expertise with NLP methods.
On the contrary, language representations play an important role here to tackle our task: we substantially explore how semantic features help characterize doctor expertise, which has not been studied before.  

Our work is also related to the previous language understanding research over doctor-patient dialogues on online health forums \cite{zeng-etal-2020-meddialog}, where various compelling applications are explored, such as information extraction \cite{ramponi-etal-2020-biomedical,du-etal-2019-extracting,zhang-etal-2020-mie}, question answering \cite{pampari-etal-2018-emrqa, xu-etal-2019-doubletransfer}, and medical report generation \cite{enarvi-etal-2020-generating}.
In comparison with them, we concern doctor expertise and characterize it from both doctor profiles and the past patient-doctor dialogues, which is a gap in previous work filled in this work.

\section{Conclusion}

This paper has studied doctor recommendation in online health forums. 
We have explored the effects of doctor profiles and history dialogues in the learning of doctor expertise through a self-learning task and a multi-head attention mechanism.
Substantial experiments on a large-scale Chinese dataset demonstrate the effectiveness of our method.

\section*{Ethical Considerations}

It should be mentioned that all data, including doctors' profiles, patients' queries, and doctor-patient dialogues, are collected from the openly accessible online health forum Chunyu Yisheng whose owners make such information visible to the public (while anonymizing patients). 
Our dataset is collected by a crawler within the constraints of the forum. 
Apart from the personal information de-identified by the forum officially, to prevent privacy leaks, we manually reviewed the collected data and deleted sensitive messages. 
Additionally, we replaced each doctor's name with a unique code randomly generated to distinguish them while protecting their privacy. We ensure there is no identifiable or offensive information in the released dataset.

The dataset, approach, and model proposed in this paper are for research purposes only and intended to facilitate studies of using NLP methods for doctor expertise learning and recommendation to allow a better user experience on online health forums. We also anticipate they could advance other NLP researches like question answering (QA) in the biomedical domain.

\section*{Acknowledgements}

This paper is substantially supported by NSFC Young Scientists Fund (62006203), a grant from the Research Grants Council of the Hong Kong Special Administrative Region, China (Project No. PolyU/25200821), PolyU internal funds (1-BE2W, 4-ZZKM, 1-ZVRH, and 1-TA27), CCF-Tencent  Open Fund (R-ZDCJ), and CCF-Baidu Open Fund (No. 2021PP15002000).
The authors would like to thank Yuji Zhang and the anonymous reviewers from ACL 2022 for their insightful suggestions on various aspects of this work.
\bibliography{custom-with_page}
\bibliographystyle{acl_natbib}

\appendix

\section{More Case Study Results}
\label{appendix:case_study}

To provide more insight of why our model can exhibit superior performance, we further discuss two more cases to understand how the multi-head attention mechanism makes use of the information from both the doctors' profiles and their history dialogues, in addition to example cases shown in \Cref{fig:intro-case} and \Cref{tab:case_att}.
Because a dialogue is mostly lengthy (as shown in Table \ref{tab:statistics}), we only show the dialogue snippets in English translations for a better display (while the model is fed with the entire dialogues in the experiments).

We present in \Cref{tab:case2_apd_texts} a case sampled from the Department of Gynecology. As can be seen, the profile of the doctor is short, while the attended dialogues provide detailed information for the symptoms, treatments, and medicine. The top 5 keywords identified by the sum of attention weights for each head are shown in \Cref{tab:case2_apd_texts}(b).While several heads seem to attend to one or two specific tokens, for example head $1$, $4$, and $5$ attend to the token ``menstruation'', we observe each head has its own focus. For example, it is reasonable to infer that head $1$ concerns messages related to the preparation of  pregnancy, head $4$ irregular period, and head $5$ prognosis of abortion.

\begin{table}[h]\small
\begin{minipage}{0.5\textwidth}
\newcommand{\tabincell}[2]{\begin{tabular}{@{}#1@{}}#2\end{tabular}}
\begin{center}
\begin{tabular}{|l|}
\hline
\tabincell{l}{\textbf{Query $q$ from Anonymous Patient $P$}\\
There is brown secreta after my last menstruation, and it\\
disappears after sex. What’s wrong with me?}\\
\hline
\hline
\tabincell{l}{\textbf{Profile $p$ of Doctor $D$}\\
30 years of experience in obstetrics and gynecology.}\\
\hline
\hline
\tabincell{l}{\textbf{Attended Dialogue $d_1$}\\
$u_P$: The urine test result for pregnancy is negative for the\\
24th day after my last sexual behaviour, and it is the same\\
for the 20th, 22nd, 23rd. Can I rule out pregnancy?\\
$u_P$: I took contraceptive pill last month. I don’t know my\\
current ovulation.\\
$u_D$: Not pregnant, don’t worry.\\
$u_P$: I've been getting yellowish vaginal discharge lately.\\
Am I inflamed?\\
$u_D$: It’s fungal vaginitis. I suggest you take fluconazole\\
pills.}\\
\hline
\tabincell{l}{\textbf{Attended Dialogue $d_2$}\\
$u_P$: My boyfriend and I had sex with condom. It was the\\
first time for me, but my boyfriend had had sex life with\\
others. Is there a high chance I get infected with HPV?\\
$u_D$: From your description, it is not likely to happen.}\\
\hline
\tabincell{l}{\textbf{Attended Dialogue $d_3$}\\
$u_P$: My period has been lasting for 8 days. I bleed a lot\\
and have blood clots. What’ the matter with me?\\
$u_P$: In the past year my period has always been regular.\\
But in the past two months I took Ejiao for a few days.\\
$u_D$: It’s abnormal and it could be caused by Ejiao. I\\
suggest hemostasis, or it could lead to anemia.}\\
\hline
\tabincell{l}{\textbf{Attended Dialogue $d_4$}\\
$u_P$: My vaginal opening is like white petal. I have had\\
sexual experience, but I feel alright. Is it condyloma\\
acuminatum?\\
$u_D$: How long has this lasted? How are you feeling?\\
$u_P$: I feel nothing.\\
$u_D$: It’s normal, not condyloma acuminatum. It is likely\\
to be hymen residue.}\\
\hline
\end{tabular}
\end{center}
\vspace{-1em}
\caption*{(a)}
\end{minipage}


\begin{minipage}{0.5\textwidth}
\newcommand{\tabincell}[2]{\begin{tabular}{@{}#1@{}}#2\end{tabular}}
\begin{center}
\begin{tabular}{|cl|}
\hline

\textbf{Head} $i$ &\textbf{Top 5 Keywords}\\

\hline
\hline

\textsc{1}
&\tabincell{l}{怀孕，生理，月经，性行为，排卵期\\
(pregnancy, physiology, menstruation, sexual\\
behaviour, ovulation)}\\
\hline

\textsc{2}
&\tabincell{l}{炎症，白带，阴道，宫颈，分泌物\\
(inflammation, leukorrhea, vagina, cervix\\
uteri, secreta)}\\
\hline

\textsc{3}
&\tabincell{l}{性行为，预防措施，避孕药，艾滋，精子\\
(sexual behaviour, precaution, contraceptive,\\
AIDS, sperm)}\\
\hline

\textsc{4}
&\tabincell{l}{月经，激素，子宫内膜，出血，避孕药\\
(menstruation, hormone, endometrium,\\
bleeding, contraceptive)}\\
\hline

\textsc{5}
&\tabincell{l}{月经，怀孕，性行为，排卵期，流产\\
(menstruation, pregnancy, sexual behaviour,\\
ovulation, abortion)}\\
\hline

\textsc{6}
&\tabincell{l}{增生，肿块，卵巢，痛经，宫颈\\
(hyperplasia, lump, ovary, dysmenorrhea,\\
cervix uteri)}\\

\hline
\end{tabular}
\end{center}
\vspace{-1em}
\caption*{(b)}
\end{minipage}

\caption{(a) The sample patient query $q$ from anonymous patient $P$ on the top, followed by the profile of a sample doctor $D$ and four dialogues $D$ engaged before. $u_P$ refers to utterances of $P$, and $u_D$ utterances of $D$. (b) The top 5 medical terms attended by each head given the input sample in \Cref{tab:case2_apd_texts}(a). The medical terms are from the THUOCL lexicon in Section \ref{sec:dataset}.}
\label{tab:case2_apd_texts}
\end{table}

Table \ref{tab:case1_apd_texts} shows another example sampled from the Department of Dermatology. In this case, the doctor's profile is more detailed while generic. Top 5 keywords for each attention head are shown in \Cref{tab:case1_apd_texts}(b). Similar to the observation from Table \ref{tab:case2_apd_texts}, the token ``pruritus'' occurs in most attended keywords of 5 heads for that it is one of the most common symptoms, whereas each head focuses on different aspects related to the query.

\begin{table}[h]\small

\begin{minipage}{0.5\textwidth}
\newcommand{\tabincell}[2]{\begin{tabular}{@{}#1@{}}#2\end{tabular}}
\begin{center}
\begin{tabular}{|l|}
\hline

\tabincell{l}{\textbf{Query $q$ from Anonymous Patient $P$}\\
In the past week, she has been keeping saying that her\\
back, her legs, and her whole body were all itchy. I\\
observe she has a few dry eczema spots on her body, a\\
little wrinkled and peeling.}\\

\hline
\hline

\tabincell{l}{\textbf{Profile $p$ of Doctor $D$}\\
Good at treating common skin diseases, including\\
diagnosis and treatment of acne, urticaria, viral warts,\\
eczema, shingles, etc.}\\

\hline
\hline

\tabincell{l}{\textbf{Attended Dialogue $d_1$}\\
$u_P$: I've had beriberi for over a year. At night, I feel itchy\\
and the skin of my feet peels off.\\
$u_D$: I suggest you apply topical antifungal ointment to\\
your feet and wash socks with boiled water every day. It\\
takes 4-6 weeks to cure tinea pedis.}\\

\hline

\tabincell{l}{\textbf{Attended Dialogue $d_2$}\\
$u_P$: It's red and itchy around my mouth and nose. What’s\\
the matter with me?\\
$u_D$: Are they blisters or pimples? You possibly got\\
seborrheic dermatitis.\\
$u_P$: My husband has beriberi, is it possible I'm infected\\
by him?\\
$u_D$: Not likely.}\\

\hline

\tabincell{l}{\textbf{Attended Dialogue $d_3$}\\
$u_P$: I froze this spot, is it going to scab and peel off?\\
$u_D$: It’s already dark red, so theoretically it should soon\\
peel off.\\
$u_P$: It’s nearly fourteen days after freeze, can I bath now?\\
$u_D$: You could shower but should not bath. Be careful not\\
to irritate this spot.}\\

\hline

\tabincell{l}{\textbf{Attended Dialogue $d_4$}\\
$u_P$: I have nail fungus, and I felt itchy after I applied\\
ciclopirox amine cream the day before yesterday. Today I\\
observe my toes swell.\\
$u_D$: There is a possible delayed allergic reaction to the\\
drug. I suggest you rinse your toes with warm water and\\
stop applying that cream.}\\

\hline
\end{tabular}
\end{center}
\vspace{-1em}
\caption*{(a)}
\end{minipage}
\begin{minipage}{0.5\textwidth}

\newcommand{\tabincell}[2]{\begin{tabular}{@{}#1@{}}#2\end{tabular}}
\begin{center}
\begin{tabular}{|cl|}
\hline

\textbf{Head} $i$ &\textbf{Top 5 Keywords}\\

\hline
\hline

\textsc{1}
&\tabincell{l}{瘙痒、湿疹、红肿、刺激、疱疹\\
(pruritus, eczema, redness and swelling,\\
irritation, herpes)}\\
\hline

\textsc{2}
&\tabincell{l}{瘙痒、脱皮、传染、皮癣、细菌\\
(pruritus, desquamation, infection, ringworm,\\
bacteria)}\\
\hline

\textsc{3}
&\tabincell{l}{过敏，红肿，皮炎，瘙痒，药膏\\
(allergy, redness and swelling, dermatitis,\\
pruritus, ointment)}\\
\hline

\textsc{4}
&\tabincell{l}{红肿，皮炎，痤疮，伤疤，粉刺\\
(redness and swelling, dermatitis, acne, scar,\\
pimple)}\\
\hline

\textsc{5}
&\tabincell{l}{脱皮，开裂，瘙痒，红斑，感染\\
(desquamation, chap, pruritus, erythema,\\
infection)}\\
\hline

\textsc{6}
&\tabincell{l}{瘙痒，性病，传染，疱疹，尖锐湿疣\\
(pruritus, venereal disease, infection, herpes,\\
condyloma acuminatum)}\\

\hline
\end{tabular}
\end{center}
\vspace{-1em}
\caption*{(b)}
\end{minipage}
\caption{(a) The sample patient query $q$  from anonymous patient $P$ on the top, followed by the profile of a sample doctor $D$ and four dialogues $D$ engaged before. $u_P$ refers to utterances of $P$, and $u_D$ utterances of $D$. (b) The top 5 medical terms attended by each head given the input sample in \Cref{tab:case1_apd_texts}(a). The medical terms are from the THUOCL lexicon in Section \ref{sec:dataset}.}
\label{tab:case1_apd_texts}
\end{table}

\end{CJK}
\end{document}